\documentclass[letterpaper, 10 pt, conference]{ieeeconf}  

\IEEEoverridecommandlockouts                              

\usepackage{graphics} 
\usepackage{epsfig} 
\usepackage{mathptmx} 
\usepackage{times} 
\usepackage{amsmath} 
\usepackage{amssymb}  
\usepackage{booktabs}
\usepackage{svg}
\usepackage{tikz}
\usepackage{pgfplots}
\usepackage{color}
\usepackage{cuted}    
\usepackage{capt-of}
\usepackage{lipsum}

\title{\LARGE \bf
RopeDreamer: A Kinematic Recurrent State Space Model for Dynamics of Flexible Deformable Linear Objects
}

\author{Tim Missal$^{\star \dagger 1}$, Lucas Domingues$^{\star2,3}$, Berk Guler$^{1,4}$, Simon Manschitz$^{4}$, 
\\Jan Peters$^{1,5-8}$, Paula Dornhofer Paro Costa$^{2,9}$
\thanks{$^\star$Equal contribution; $^\dagger$ Part of this work was performed during an exchange at UNICAMP $^1$Technical University of Darmstadt $^2$School of Electrical and Computer Engineering, Universidade Estadual de Campinas (UNICAMP), Brazil $^3$Instituto de Pesquisas Eldorado, Brazil $^4$Honda Research Institute Europe GmbH $^5$German Research Center for Artificial Intelligence (DFKI) $^6$hessian.AI $^7$Robotics Institute Germany (RIG) $^8$Centre for Cognitive Science $^9$Artificial Ingelligence Lab, Recod.ai; Corresponding authors: \texttt{tim.missal@stud.tu-darmstadt.de}, \texttt{lucas.domingues@eldorado.org.br}}
\thanks{This work was partially funded by the Coordenação de Aperfeiçoamento de Pessoal de Nível Superior – Brasil (CAPES) – Finance Code 001, by the São Paulo Research Foundation (FAPESP) under grant \#2020/09838-0 (BI0S - Brazilian Institute of Data Science), and by the Ministry of Science, Technology, and Innovations, with resources from Law No. 8.248, of October 23, 1991, under the PPI-SOFTEX program, DOU 01245.003479/2024-10.}
}

\begin{document}

\maketitle
\thispagestyle{empty}
\pagestyle{empty}

\begin{strip}
        \vspace{-10em}
    \centering
    \includegraphics[width=\textwidth, height=7.5cm]{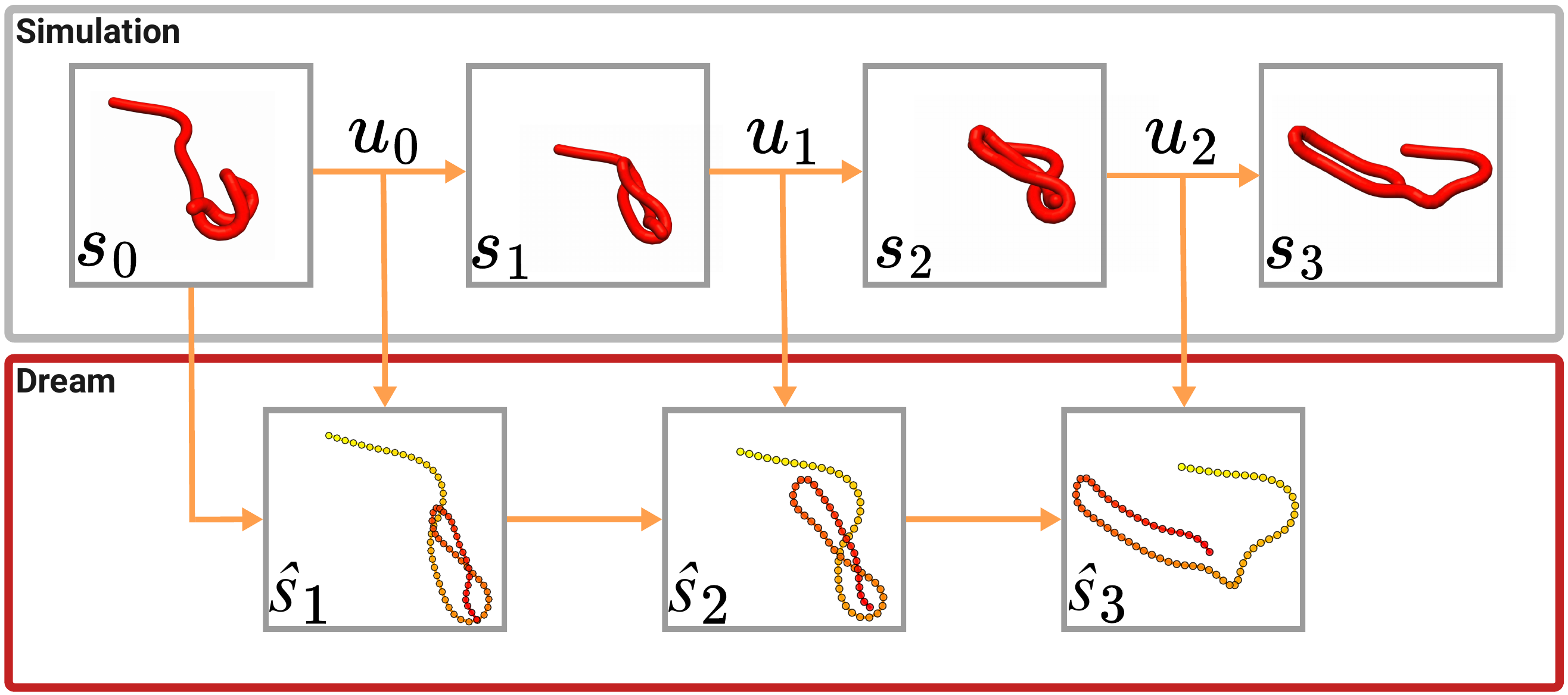}
    \captionof{figure}{After observing an initial state $s_0$ from simulation, RopeDreamer predicts the following evolution of the Deformable Linear Object given a sequence of actions $u_{0:T}$.}
    \label{fig:hero}
        \vspace{-1em}
\end{strip}
\begin{abstract}

The robotic manipulation of Deformable Linear Objects (DLOs) is a fundamental challenge due to the high-dimensional, non-linear dynamics of flexible structures and the complexity of maintaining topological integrity during contact-rich tasks. While recent data-driven methods have utilized Recurrent and Graph Neural Networks for dynamics modeling, they often struggle with self-intersections and non-physical deformations, such as tangling and link stretching. In this paper, we propose a latent dynamics framework that combines a Recurrent State Space Model with a Quaternionic Kinematic Chain representation to enable robust, long-term forecasting of DLO states. By encoding the DLO as a sequence of relative rotations (quaternions) rather than independent Cartesian positions, we inherently constrain the model to a physically valid manifold that preserves link-length constancy. Furthermore, we introduce a dual-decoder architecture that decouples state reconstruction from future-state prediction, forcing the latent space to capture the underlying physics of deformation. We evaluate our approach on a large-scale simulated dataset of complex pick-and-place trajectories involving self-intersections. Our results demonstrate that the proposed model achieves a 40.52\% reduction in open-loop prediction error over 50-step horizons compared to the state-of-the-art baseline, while reducing inference time by 31.17\%. Our model further maintains superior topological consistency in scenarios with multiple crossings, proving its efficacy as a compositional primitive for long-horizon manipulation planning. Code, models and dataset will be released in a future revision.

\end{abstract}

\section{INTRODUCTION}
\label{sec:intro}

Robotic manipulation of Deformable Linear Objects (DLOs), such as cables, ropes, and sutures, remains one of the most challenging frontiers in autonomous manipulation while having a wide range of applications, such as Cable Routing ~\cite{waltersson_planning_2022, wilson_cable_2023}  Knot Tangling ~\cite{sudry_hierarchical_2023} and Untangling ~\cite{grannen_untangling_2021, guler_towards_2025}. Unlike rigid bodies, DLOs possess an infinite-dimensional state space and exhibit complex, non-linear dynamics go\-verned by internal material constraints, self-intersections, and environmental interactions like friction. Predicting the evolution of a DLO's shape during a manipulation task is critical for planning. Yet, it requires a model that is both computationally efficient for real-time use and robust enough to maintain topological integrity over long horizons.

Recent advances in learning-based dynamics have shifted from analytical physical engines to data-driven models. Among these, Graph Neural Networks with segment-level encoders, such as GA-Net ~\cite{gu_learning_2025} have shown promise by treating the DLO as a series of interacting nodes. However, these methods often suffer from two primary drawbacks: first, they frequently struggle with over-smoothing ~\cite{oono_graph_2021} and over-squashing ~\cite{giovanni_how_2024, alon_bottleneck_2021} during long-range temporal dependencies, where the effect of an action at one end of the DLO fails to propagate accurately across many segments. Second, segment-level encoders lack inherent physical constraints, leading to non-physical ``stretching'' or ``clipping'' when the DLO undergoes complex deformations or self-intersections.

In this paper, we propose a novel latent dynamics framework for DLO manipulation that addresses these limitations by combining structured kinematic priors with state-of-the-art world models. We leverage the Recurrent State Space Model (RSSM) ~\cite{hafner_dream_2020} to project DLO states into a latent manifold, effectively separating deterministic temporal dependencies from stochastic environmental uncertainties.

To ensure physical consistency, we depart from standard Cartesian position representations and introduce a Quaternionic Kinematic Chain. By modeling the DLO as a sequence of unit quaternions representing relative rotations between equidistant segments, we inherently constrain the state space to a valid manifold, preventing non-physical stretching by design. Furthermore, we introduce a dual-decoder strategy that separates state reconstruction from predictive forecas\-ting, ensuring that the model remains grounded in the current observation while optimizing the latent space specifically for multi-step ``dreaming'' of future DLO configurations.

Our contributions are three-fold:
\begin{enumerate}
\item We adapt the RSSM framework for high-segment DLO dynamics, enabling stable open-loop predictions over long horizons without ground-truth correction.
\item We introduce a quaternionic state representation that enforces link-length constancy and provides superior topological consistency during complex maneuvers involving self-intersections.
\item We demonstrate through extensive simulation\footnote{The dataset, models, and evaluation scripts are  available at: to be released} that our approach outperforms current baselines in long-horizon accuracy and maintains the DLOs topology better, while reducing inference latency by at least 31.17\%.
\end{enumerate}

The remainder of this paper is organized as follows: Section \ref{sec:problem} formalizes the DLO dynamics problem; Section \ref{sec:method} details our RSSM-based architecture and quaternionic representation; Section \ref{sec:results} presents quantitative results against a baseline model; and concludes with a discussion.

\section{Related Work}
\label{sec:related}

Robotic manipulation of DLOs spans a broad set of tasks and approaches. In this section, we review shape matching tasks and the dynamics prediction of DLOs.

\subsection{DLO Shape Matching}

Existing research on robotic shape matching of DLOs can be categorized into physical properties and manipulation settings. A common distinction is between rope-like DLOs that behave as highly flexible objects and cable-like DLOs whose internal stiffness strongly constrains deformation. Research on ropes focuses on flexible DLOs with negligible bending stiffness ~\cite{yan_learning_2020, zhang_deformable_2021, lee_sample-efficient_2022, yan_self-supervised_2020}. These objects are typically manipulated on a planar surface, where they maintain their state through friction and are unconstrained, thus allowing for manipulation at any point along their length. Ropes are challenging to work with because of their non-linear dynamics and their diverse set of possible states. In contrast, research on cables focuses on DLOs characterized by significant internal stiffness. As cables do not maintain their shape solely through surface friction, the problem statements involving cables generally assume boundary conditions where both ends of the DLO are secured, either to a rigid fixture or a robotic end-effector ~\cite{gu_learning_2025, yue_lstm-gcn_2025, yu_hybrid_2025, yu_shape_2022, yang_learning_2021}. This work specifically addresses the challenges of manipulating unconstrained, flexible DLOs.

\subsection{Data-Driven DLO Dynamics Prediction}

The data-driven dynamics models utilized in DLO mani\-pulation can be further classified by their state representation. One approach typically followed in rope manipulation employs latent visual models, which predict changes in the DLO state by encoding visual input into a latent space and predicting visual observations of following timesteps ~\cite{yan_learning_2020, zhang_deformable_2021, lee_sample-efficient_2022, li_deformnet_2024}. However, these latent models frequently struggle to correctly interpret self-intersecting rope structures, such as loops or knots, while state-estimation frameworks such as TrackDLO ~\cite{xiang_trackdlo_2023} or other image-based methods ~\cite{choi_mbest_2023, caporali_rt-dlo_2023} demonstrate greater robustness in such scenarios. It has further been shown that predicting dynamics from pixels is outperformed by direct state representation, highlighting the need to split DLO Tracking and Dynamics Modeling ~\cite{yan_self-supervised_2020}. 

Alternatively, other models represent the DLO explicitly as a sequence of discrete segments ~\cite{yan_self-supervised_2020, gu_learning_2025, yue_lstm-gcn_2025, yu_hybrid_2025, yu_shape_2022, yang_learning_2021}. These state-based models are further divided into recurrent architectures, such as Long Short-Term Memory (LSTM) networks ~\cite{yan_self-supervised_2020, yang_learning_2021}, and structural representations utilizing Graph Neural Networks (GNNs) or hybrids of both ~\cite{yue_lstm-gcn_2025, yu_hybrid_2025, yu_shape_2022}. While showing promise in handling constrained, short DLOs, these methods are limited by self-intersections (e.g. a segment at one end of the DLO affects a segment in the middle of it), as local message-passing is fundamentally unfit to handle long-range dependencies ~\cite{alon_bottleneck_2021}. Although some approaches attempt to mitigate this by incorporating attention mechanisms, these techniques still face challenges; for instance, in a non-intersected DLO configuration, the state of a segment typically does not affect distant segments, which can make attention weights unreliable in such situations ~\cite{mudarisov_limitations_2025}. Furthermore, attention mechanisms scale poorly ($O(N^2)$) to large sequence lengths ~\cite{patro_mamba-360_2024}.

Another approach to learn dynamics of a DLO from data is to approximate the Jacobian Matrix of the DLO locally ~\cite{yu_global_2023, jin_robust_2019}. While making efficient use of sparse data, they are limited to small deformations ~\cite{gu_survey_2023}. RSSMs have further shown promising results in modeling deformables ~\cite{li_deformnet_2024}. 

Our approach synthesizes key strengths from these previous works. We represent the DLO state as coordinates in $\mathbb{R}^3$, but rather than leveraging GNNs to model local graph-based connectivity, we treat the DLO as a kinematic chain. While prior methods rely on local neighbor interactions, attention mechanisms, or LSTMs to capture segment-wise dependencies, we utilize an RSSM to encode the global state into a latent manifold. The RSSM captures the underlying system dynamics in a latent space that is split into a deterministic and a stochastic part, which is better suited for modeling long-term temporal dependencies and maintaining physical consistency over extended horizons than purely deterministic approaches ~\cite{hafner_learning_2019, hafner_dream_2020}.

\section{Problem Formulation}
\label{sec:problem}

\begin{figure}[t]
    \centering
    \includegraphics[width=\linewidth]{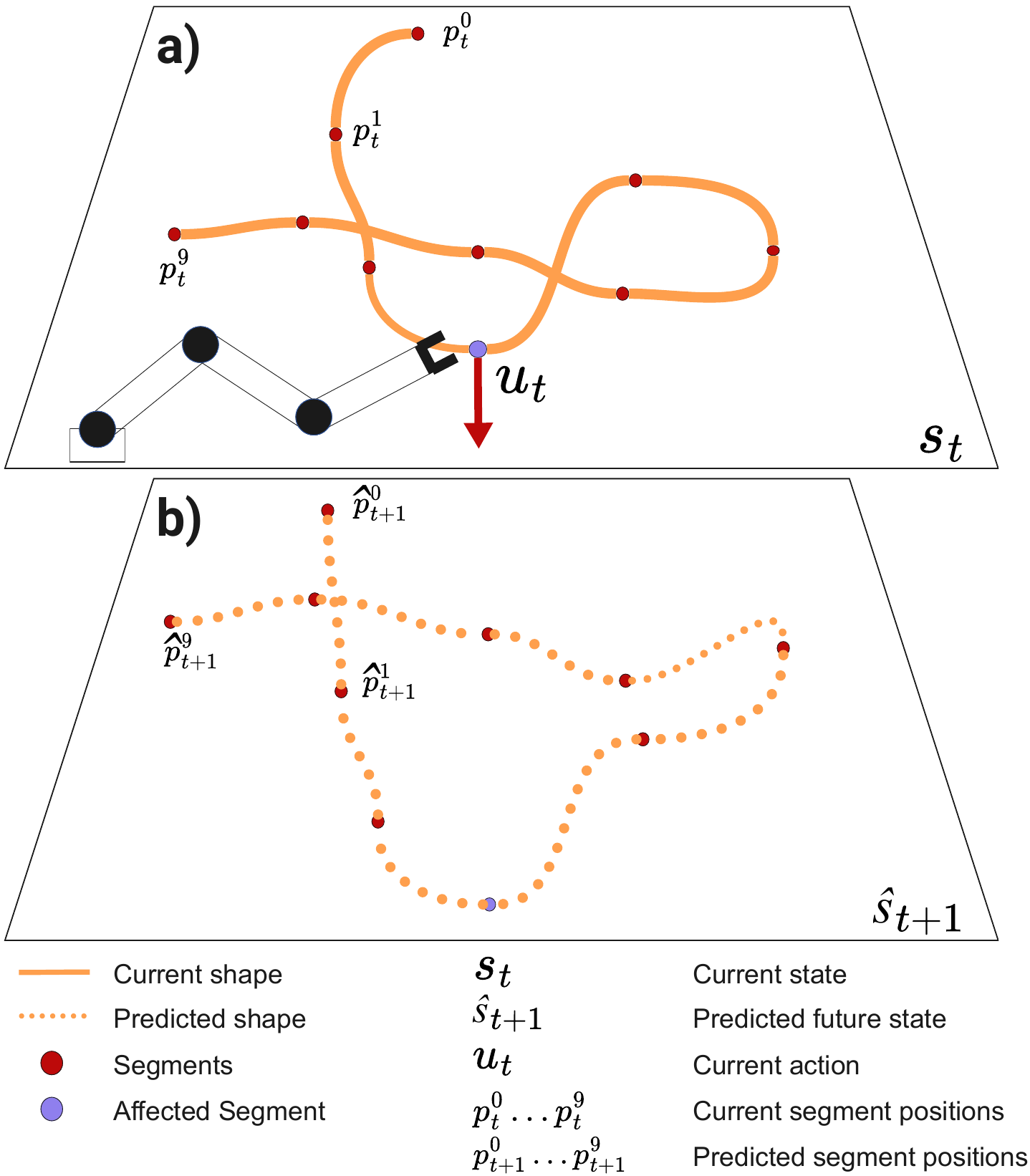}
    \caption{Problem formulation of DLO shape prediction task. a) The DLO lies on a work plane without fixation and can be moved by the end-effector at an arbitrary point. Overlaps between rope segments are thus possible. b) The affected segment is moved downwards, affecting the entirety of the DLO.}
    \label{fig:setup}
        \vspace{-1.5em}
\end{figure}

We consider the task of modeling the dynamics of a highly flexible DLO during planar manipulation by a robotic end-effector, as depicted in Fig. \ref{fig:setup}. The DLO is assumed to have low bending stiffness, allowing for a wide range of DLO states. We treat the DLO as a free-moving multi-body system consisting of L segments, where each segment lives in $\mathbb{R}^3$. The system's behavior is governed by internal bending constraints, friction with the ground plane, and interactions with a robotic end effector.

\subsection{State and Action Representation}
The DLO is discretized into a sequence of $L$ equidistant segments. The state of the DLO at timestep $t$ is denoted as $\mathbf{s}_t \in \mathbb{R}^{L \times 3}$, representing the positions of all $L$ segments:
\begin{equation*}
    \mathbf{s}_t = [\mathbf{p}^0_{t}, \mathbf{p}^1_{t}, \dots, \mathbf{p}^{L-1}_{t}]
\end{equation*}

where $p_i \in \mathbb{R}^3$ is the position of the $i$-th segment. 

The robotic interaction is defined by a pick-and-place action $u_t = (i_g, \Delta p)$, where $i_g \in \{1, \dots, L\}$ denotes the index of the specific segment grasped by the robotic end-effector and $\Delta p = [\Delta x, \Delta y]^\top \in \mathbb{R}^2$ represents the relative displacement vector of the grasped segment during the translation phase of the pick-and-place trajectory. This action consists of a grasp at an arbitrary point along the DLO's length, followed by a trajectory comprising a vertical lift along the Z axis, a translation in the $XY$-plane, and a subsequent vertical descent along the Z axis to the support surface.

\subsection{Objective}
The goal of this work is to develop a dynamics model $f_\theta$ capable of accurately predicting future DLO states in 3D-space given the current state, a historical context, and an intended action. Given the state $s_t$, a history of $n$ previous states $s_{t-n:t-1}$, and an action $u$, the model seeks to predict the subsequent state $\hat{s}_{t+1} = f_\theta(s_{t-n:t}, u_t)$.

By formulating the problem around relative displacements on the $XY$-plane and making the assumption that the DLO has already been picked up at the time of planning, the resulting model can be used to predict various displacement lengths without retraining by simply splitting up large actions into smaller ones. This incremental approach allows the model to generalize to large-scale deformations by iteratively compounding small-scale displacements.

\subsection{Evaluation Metric}
To quantify the accuracy of the dynamics model, we employ the Root Mean Square Error (RMSE) between the predicted segment positions $\hat{s}_{t+1}$ and the ground truth positions $s_{t+1}$. The RMSE for a DLO state is defined as
\begin{equation*}
    \text{RMSE} = \sqrt{\frac{1}{L} \sum_{i=1}^{L} \|\mathbf{p}^i_{t+1} - \hat{\mathbf{p}}^i_{t+1}\|^2}
\end{equation*}
where $p_{i}$ and $\hat{p}_{i}$ are the ground truth and predicted coordinates for the $i$-th segment, respectively.

\section{Dynamics Model for Deformable Linear Objects} 
\label{sec:method}

In this section, we describe our proposed latent dynamics framework for modeling the transitions of DLOs.

\begin{figure*}[t] 
    \centering
    \includegraphics[width=\textwidth]{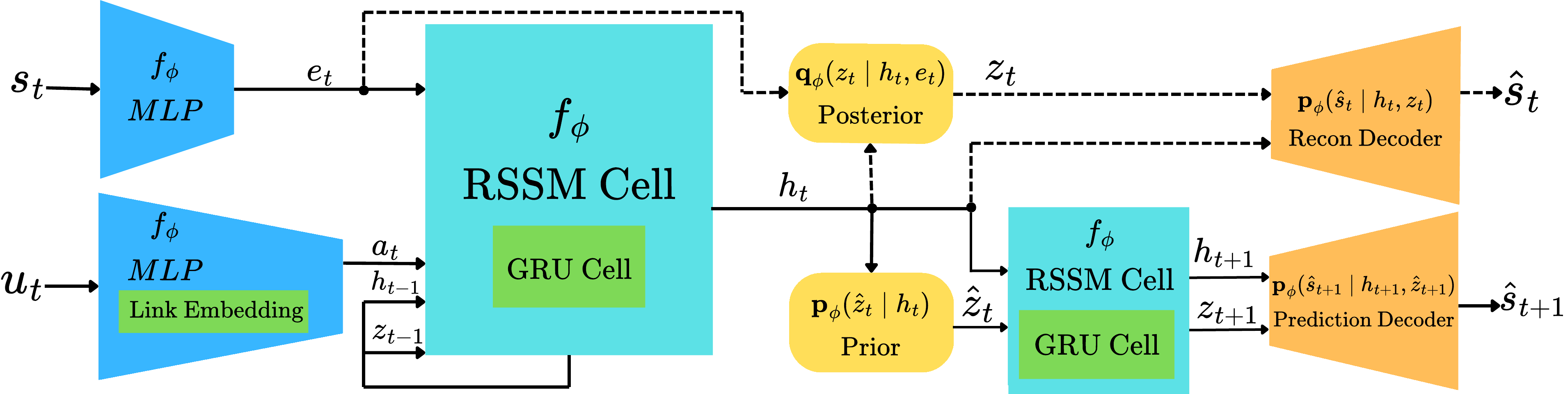}
    \caption{Architecture of the World Model for DLO manipulation. The current DLO state and manipulation actions are first fed into their respective encoders to generate latent embeddings. These representations are passed to the Recurrent State Space Model (RSSM) cell, alongside the previous deterministic and stochastic states, to update the current deterministic memory ($h_t$) and compute both the prior ($\hat{z}_t$) and posterior ($z_t$) distributions. In our dual-decoder setup, the posterior and deterministic states are used by the Reconstruction Decoder to reconstruct the current observation for spatial grounding. Simultaneously, the prior is chained through a subsequent RSSM step to forecast future latent variables, which the Prediction Decoder uses to predict the next physical rope state. Dashed lines indicate pathways and components utilized exclusively during the training phase.}
    \label{fig:dreamer_rope_arc}
    \vspace{-0.5cm}
\end{figure*}

\subsection{Quaternionic Kinematic Chain}
To treat the DLO as a rigid-link structure, as part of pre-processing, we transform the state $s_t$ into a hybrid representation. The first segment of the DLO is represented by its position $p^0 \in \mathbb{R}^3$, while the configuration of all subsequent segments is defined by unit quaternions $q^i \in \mathbb{H}$ representing the relative rotation between link~$i-1$ and link~$i$. 

Assuming a fixed link length, the position of any segment position ~$p^i$ can be reconstructed via forward kinematics. This formulation constrains the state-space to a valid mani\-fold, inherently preventing the model from predicting non-physical stretching. Furthermore, as the model is solely presented with the DLO's geometry and does not have to encode the distance between segments as in previous approaches, this pre-processing allows to scale a pretrained model to DLOs of varying length. The full state vector 
\begin{equation*}
    \mathbf{s}_t = [\mathbf{p}^0_{t}, \mathbf{q}^1_{t}, \mathbf{q}^2_{t}, \dots, \mathbf{q}^{L-1}_{t}]^\top \in \mathbb{R}^{(3+4(L-1)) \times 1}
\end{equation*}  
is derived by concatenating the position of the base $p^0$ and the relative rotations of all subsequent segments in quaternion form $q^{1:L-1}$. 

\subsection{Latent Dynamics Learning using RSSM}

Our objective is to train a dynamics model capable of predicting the evolving state $s_{t+1}$ given a history of past states and control actions $u_t$. Standard sequential predictors (e.g. LSTMs) often struggle with the information washout inherent in high-segment DLOs, where local perturbations propagate complexly across the entire structure. GNNs suffer from over-smoothing ~\cite{oono_graph_2021} and over-squashing ~\cite{giovanni_how_2024, alon_bottleneck_2021}. Attention mechanisms only partially solve these problems for self-intersecting DLOs and their computational complexity grows exponentially with sequence length ~\cite{patro_mamba-360_2024}.

To mitigate these long-range dependencies, we propose predicting dynamics within a latent state-space using an RSSM ~\cite{hafner_learning_2019}.

The RSSM operates directly on latent representations by first projecting inputs into embeddings via an encoder. It then decomposes the model's dynamics into distinct deterministic and stochastic components. The deterministic state $h_t$ (maintained by a Gated Recurrent Unit (GRU)) captures long-term memory and temporal dependencies, while the stochastic state $z_t$ captures the uncertainty and variability of the environment that a single vector cannot represent ~\cite{hafner_learning_2019}. During training, the model uses a \textit{Posterior} distribution grounded in actual observations to learn the latent space, while simultaneously training a \textit{Prior} (Transition Predictor) to forecast the next latent state without seeing the future, optimizing the model to keep these distributions similar to each other. This allows the agent to ``dream'' or simulate long sequences entirely in the latent space by chaining the prior and the recurrent model ~\cite{hafner_dream_2020}.

Unlike existing latent models that typically operate on high-dimensional raw visual input, our approach leverages an explicit kinematic representation, available via ground-truth data in simulation or through state-of-the-art tracking algorithms such as TrackDLO \cite{xiang_trackdlo_2023} in real-world settings. This decoupling allows the RSSM to focus specifically on the highly non-linear transitions of DLO dynamics while maintaining the flexibility to incorporate various perception backbones. Given that DLO tracking is a standalone research challenge, our framework is designed to be modular, ensuring compatibility with evolving state-estimation methods. By integrating this structured representation with a world-model framework, we address the gap in scaling latent dynamics to high-segment, potentially overlapping DLOs, a domain currently underexplored in frameworks using segment-level encoding like GA-Net ~\cite{gu_learning_2025}. 
The main components of our RSSM are:

\begin{tabbing}
\textbf{Reconstruction Decoder:} \qquad \= \kill 
\textbf{Action Encoder:} \> $a_t = f_\phi(\text{u}_t)$ \\
\textbf{Recurrent Model:} \> $h_t = f_\phi(h_{t-1}, z_{t-1}, a_{t-1})$ \\
\textbf{State Encoder:} \> $e_t = f_\phi(s_t)$ \\
\textbf{Posterior:} \> $z_t \sim q_\phi(z_t \mid h_t, e_t)$ \\
\textbf{Prior:} \> $\hat{z}_t \sim p_\phi(\hat{z}_t \mid h_t)$ \\
\textbf{Reconstruction Decoder:} \> $\hat{s}_t^{\text{recon}} \sim p_\phi(\hat{s}_t \mid h_t, z_t)$ \\
\textbf{Prediction Decoder:} \> $\hat{s}_{t+1}^{\text{pred}} \sim p_\phi(\hat{s}_{t+1} \mid h_{t+1}, \hat{z}_{t+1})$
\end{tabbing}

The key aspects of this RSSM implementation designed to fulfill the supervised objective of predicting the next state rely on the following points:

\subsubsection{Link-Aware Action Encoding}
The model employs a specialized \textbf{Action encoder}. It uses a learnable embedding layer for the DLO link index and a separate MLP for the link displacement. This allows the model to learn specific physical dynamics for different segments of the rope, capturing the non-homogeneous behavior often seen in DLO manipulation.

\subsubsection{Dual-Decoder Strategy}
The architecture splits the decoding task into a \textbf{Reconstruction Decoder} and a \textbf{Prediction Decoder}. The reconstruction decoder acts similarly to a traditional self-supervised auto-encoder ~\cite{hinton_reducing_2006}, grounding the latent representations in the current state's geometry and position. Conversely, the prediction decoder explicitly forecasts the next state $s_{t+1}$ based on the prior ``dream'' step. Separating these tasks prevents the reconstruction loss from dominating the predictive dynamics, allowing the model to learn a transition space optimized for long-horizon DLO deformation forecasting.

\subsubsection{Optimization and Loss}
The model's optimization is rooted in maximizing the Evidence Lower Bound (ELBO), a fundamental concept derived from variational inference ~\cite{kingma_auto-encoding_2022}. ELBO is widely used in domains where the true underlying probability distribution is computationally intractable to calculate directly. Fundamentally, it balances two competing objectives: it forces the model to accurately reconstruct the observed data while simultaneously keeping the learned latent representations organized and close to a known prior distribution. By leveraging the ELBO, the RSSM is compelled to learn a compressed, probabilistic latent space that effectively captures the uncertainty of the DLO's physical deformations. 

To adapt this framework specifically for multi-step DLO forecasting, we modify the standard ELBO objective into a composite loss function ~\cite{kingma_auto-encoding_2022}. Instead of relying on a single observation target, our architecture minimizes the two distinct decoder losses alongside the latent regularization:

$$\mathcal{L}_{total} = \mathcal{L}_{recon} + \mathcal{L}_{pred} + \beta \mathcal{L}_{KL}$$ 

\textbf{Reconstruction Loss}: Penalizes errors in recovering the current state $s_t$ from the posterior latent distribution, ensuring that the embedding captures the DLO's immediate geometry.
    $$\mathcal{L}_{recon} = \mathbb{E}_{q_\phi(z_t \mid h_t, e_t)} [ \| \hat{s}_t^{\text{recon}} - s_t \|^2 ]$$

\textbf{Predictive Loss}: Forces the transition dynamics to be consistent with the physical evolution of the DLO by predicting the subsequent state.
    $$\mathcal{L}_{pred} = \mathbb{E}_{p_\phi(\hat{z}_{t+1} \mid h_{t+1})} [ \| \hat{s}_{t+1}^{\text{pred}} - s_{t+1} \|^2 ]$$

\textbf{KL Regularization}: Minimizes the divergence between the representation model (posterior) and the transition predictor (prior). This prevents the posterior from straying too far from the prior, ensuring the latent space remains stable and predictable when the model ``dreams'' future trajectories.
    $$\mathcal{L}_{KL} = \text{KL}[q_\phi(z_t \mid h_t, e_t) \| p_\phi(z_t \mid h_t)]$$

A diagram containing a summary of the model architecture can be seen in Fig. \ref{fig:dreamer_rope_arc}.



\section{Results and Discussion}
\label{sec:results}

We evaluate our model and demonstrate that our architecture can accurately predict unconstrained DLO states, even under self-intersection, while maintaining the topology of the DLO. We further show that our model outperforms the strong learning-based baselines GA-Net ~\cite{gu_learning_2025} and IN-BiLSTM \cite{yang_learning_2021} in long-horizon accuracy, topology fidelity and inference time. We choose to evaluate our method against GA-Net as it outperformed the Interaction Network (IN) \cite{battaglia_interaction_2016}, PropNet \cite{li_propagation_2019} and the IN-BiLSTM \cite{yang_learning_2021} in the long-horizon DLO prediction task, thus we consider it state-of-the-art although the IN-BiLSTM was not outperformed by a significant margin. We further evaluate against IN-BiLSTM to include a fundamentally different architectures than the combination of a Transformer-Encoder and an Attention Network used in GA-Net in our evaluation. Despite showing promising results, we decide not to compare our work against EA-PE-GAT \cite{yu_hybrid_2025} as it incorporates forces affecting the DLO. This strictly requires the DLO to be fixed on at least one end to sense force, making it an unfitting candidate for comparison. 

Empirical evaluation across our models hyperparameters led to the selection of three model scales (Small, Medium, Large). The variants were chosen based on their optimal tradeoff between RMSE and parameter count. The respective and total parameter counts can be seen in Table \ref{tab:model_sizes}.

\begin{table}[ht]
\vspace{-0.2cm}
\centering
\caption{Configuration of our models' hyperparameters}
\label{tab:model_sizes}
\begin{tabular}{l c c c c c r}
\toprule
\textbf{Model} & $d_{embed}$ & $d_{action}$ & $d_{rnn}$ & $d_{z}$ & $d_{hidden}$ & \textbf{Params} \\
\midrule
Small (S) & 1024 & 1024 & 512 & 64 & 512  & 7.85M  \\
Medium (M) & 1024 & 1024 & 512 & 64 & 1024  & 15.83M  \\
Large (L) & 2048 & 512 & 1024 & 64  & 2048 & 47.86M  \\
\bottomrule
\end{tabular}
\vspace{-0.5cm}
\end{table}

\subsection{Experiment Setup}

The proposed method is evaluated on a simulated dataset containing 10,000 DLO trajectories, with a fixed horizon of 100 steps per trajectory, resulting in a dataset of 1 million transitions. The dataset is divided into training (80\%), validation (10\%), and test (10\%) subsets. The simulation is implemented in MuJoCo 3.3.7 ~\cite{todorov_mujoco_2012}, where the DLO is modeled as a chain of 70 capsules with a length of $10mm$ and a thickness of $10mm$, connected by ball joints. The capsules' friction is set to $0.8$, bending stiffness of the joints is set to $0.005$. The chain lies on a ground plane with ground friction set to $1.0$ and damping set to $0.05$. For each trajectory, the DLO is initialized by picking up a random segment and lifting it $50mm$ along the Z axis to allow the following displacements to produce crossings. The resting position after initial lift denotes $s_0$. We then sample and execute a sequence of 100 random actions $u_t$. Each individual action is defined by a $50mm$ translation in the $XY$-plane, with the heading angle sampled randomly from a uniform distribution $[0, 2\pi)$. The action is executed via a mocap object welded to the targeted segment. Each resulting trajectory consists of $T+1$ states and $T$ actions. After the final action, the segment is returned to the planar surface.

\begin{table}[t]
\centering
\caption{Hyperparameter Configurations of the baselines. GA-Net uses a Transformer Encoder and Attention Network, while IN-BiLSTM uses an Interaction Network and an LSTM to model long range dependencies. As a consequence, the configurable parameters of the IN-BiLSTM are a subset of GA-Net's. XS, S, M, L, XL denote sizes XSmall, Small, Medium, Large, XLarge.}
\label{tab:model_sizes_ganet}
\setlength{\tabcolsep}{3pt} 
\begin{tabular}{l c c c c r}
\toprule
\textbf{Model} & $d_{model}$ & $d_{hidden}$ & num\_heads & num\_layers & \textbf{Params} \\
\midrule
GA-Net XS & 144 & 150 & 12 & 6 & 2.45M \\
GA-Net S & 288 & 150 & 12 & 6 & 5.83M \\
GA-Net M & 432 & 150 & 12 & 6 & 10.21M \\
GA-Net L1 & 720 & 150 & 12 & 6 & 21.94M \\
GA-Net L2 & 432 & 2048 & 12 & 6 & 26.86M \\
GA-Net XL & 1440 & 150 & 12 & 6 & 68.69M\\
\midrule
IN-BiLSTM S & - & 512 & - & 3 & 18.64M \\
IN-BiLSTM M & - & 512 & - & 5 & 31.28M \\
IN-BiLSTM L & - & 1024 & - & 3 & 74.52M  \\
\bottomrule
\end{tabular}
\vspace{-0.5cm}
\end{table}

The proposed models and all baselines were optimized for up to 200 epochs using a learning rate of $1 \times 10^{-4}$, $\beta$ set to 1 and a batch size of 32, with model checkpointing ensuring the selection of the best-performing iteration on the validation set. Early Stopping was set to 10 epochs without performance increase on the validation set. In practice, most configurations reached convergence near epoch 50. While the GA-Net (XS) baseline maintains the original layer depth and attention head count, the larger variants were derived by scaling both the embedding dimension $d_{model}$ and the hidden layer width $d_{hidden}$ to increase model capacity. We did not find that increasing layer depth and head count over the recommended values increased accuracy, For the IN-BiLSTM \cite{yang_learning_2021}, we followed the recommended number of LSTM-Layers for two of the models, while increasing hidden state size to respect the higher complexity of our problem compared to the original source. To ensure that the models aren't limited by layer count, we also increased this hyperparameter. The model configurations are shown in Table \ref{tab:model_sizes_ganet}. Training sessions were conducted on an Nvidia RTX Pro 6000 Blackwell Series. 

\subsection{Prediction Error}

We evaluate the predictive performance of our models on a long-horizon, open-loop task using the previously simulated dataset's testing subset. 

To ground our models' latent state $h$, we perform a warmup phase of 5 steps, where the model is provided with ground-truth state-action pairs $(s_t, u_t)$. Subsequently, we assess the models' ability to predict future states given only an action for 50 consecutive timesteps. During this phase, the model receives only the sequence of future actions $\{u_t\}_{t=5}^{54}$ and must iteratively predict the subsequent states $\{\hat{s}_{t+1}\}_{t=5}^{54}$ without further ground-truth corrections. We conduct 500 independent rollouts and report the average RMSE standard deviation over timesteps $t$. 

\begin{figure}[htbp]
    \centering
    \includegraphics[width=\columnwidth]{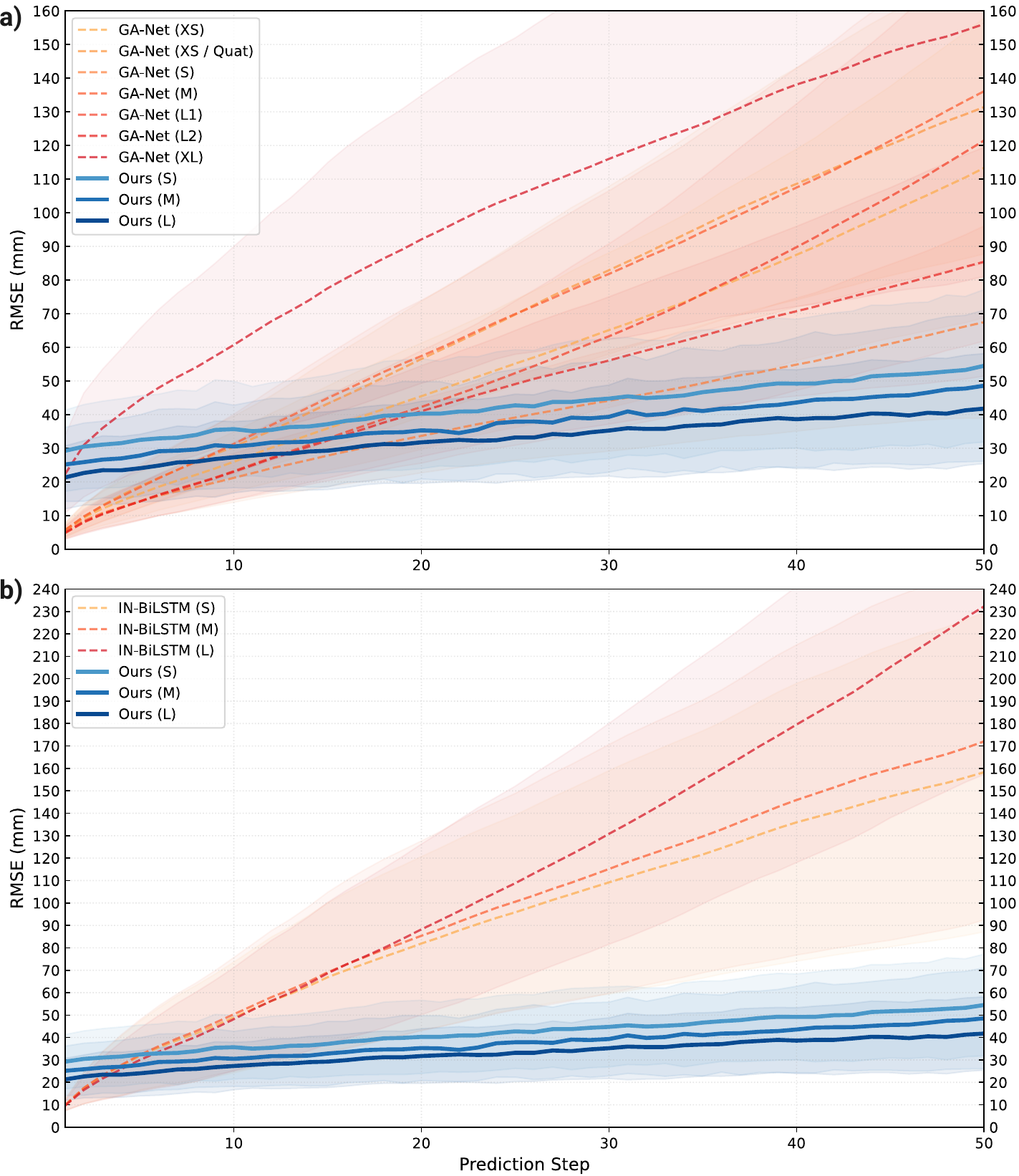}
    \caption{Comparison of RopeDreamer variants (Ours) and baseline variants on the open-loop RMSE development over 50 steps. Lower is better. (a) Ours vs. GA-Net, (b) Ours vs. IN-BiLSTM}
    \label{fig:rmse_plot}
    \vspace{-0.5cm}
\end{figure}

To evaluate the model's capacity for simultaneous state estimation, we conduct an experiment requiring the joint prediction of the DLO's shape and its global translation. Quantitative results for this prediction task are reported in Fig. \ref{fig:rmse_plot}, which illustrates the RMSE across the open-loop horizon.

Analysis of the data reveals a fundamental trade-off between local reconstruction precision and long-term dynamical stability. As depicted in Fig. \ref{fig:rmse_plot}a, most GA-Net configurations demonstrate higher accuracy in the immediate short-term, suggesting that its per-segment encoding is effective at preserving local geometric information during the initial state transitions. However, this local precision does not reliably translate to long-term stability. The sharp error growth observed, increasing by \textbf{15.68mm} at $t=10$ from $t=1$ and reaching \textbf{64.94mm} by $t=50$ for the best baseline (S), highlights a failure to model the cumulative global effects of actions over time. The IN-BiLSTM (Fig. \ref{fig:rmse_plot}b) mirrors the downward trend observed in GA-Net, albeit with a more accelerated rate of RMSE increase across the prediction horizon. This suggests that while the purely deterministic nature is accurate for single-step transitions, it suffers from significant drift during long-horizon propagation. The combination of the Transformer Encoder and an Attention Network used in GA-Net seems superior to the Interaction Network paired with a Sequential Model used in IN-BiLSTM.

In contrast, our model begins with a higher RMSE than GA-Net, which can be attributed to the information bottleneck inherent in encoding the entire DLO configuration into a latent manifold. Following this initial reconstruction penalty, our architecture exhibits stable error growth, with an increase of only \textbf{5.44mm} at $t=10$ and \textbf{19.05mm} at $t=50$ for our best model (Large). Our model further shows stable standard deviation over all timesteps while the baselines follows an exponential trend. This stability indicates that the RSSM framework effectively internalizes the underlying physics of the actions, allowing the model to maintain the global structure of the DLO even as errors in the segment-based baseline begin to diverge. At $t=50$, our best model decreases RMSE by $\mathbf{40.52\%}$ when compared against the best baseline model. 

Our ablation study using the quaternionic representation within the GA-Net framework (GA-Net XS / Quat) further clarifies these findings. While the quaternionic constraints improve GA-Net's short-term consistency, they fail to arrest the long-term divergence. This suggests that the predictive stability of our approach is primarily driven by the RSSM's latent temporal modeling rather than the coordinate representation alone.

\begin{figure}[t]
    \centering
    \includegraphics[width=\columnwidth]{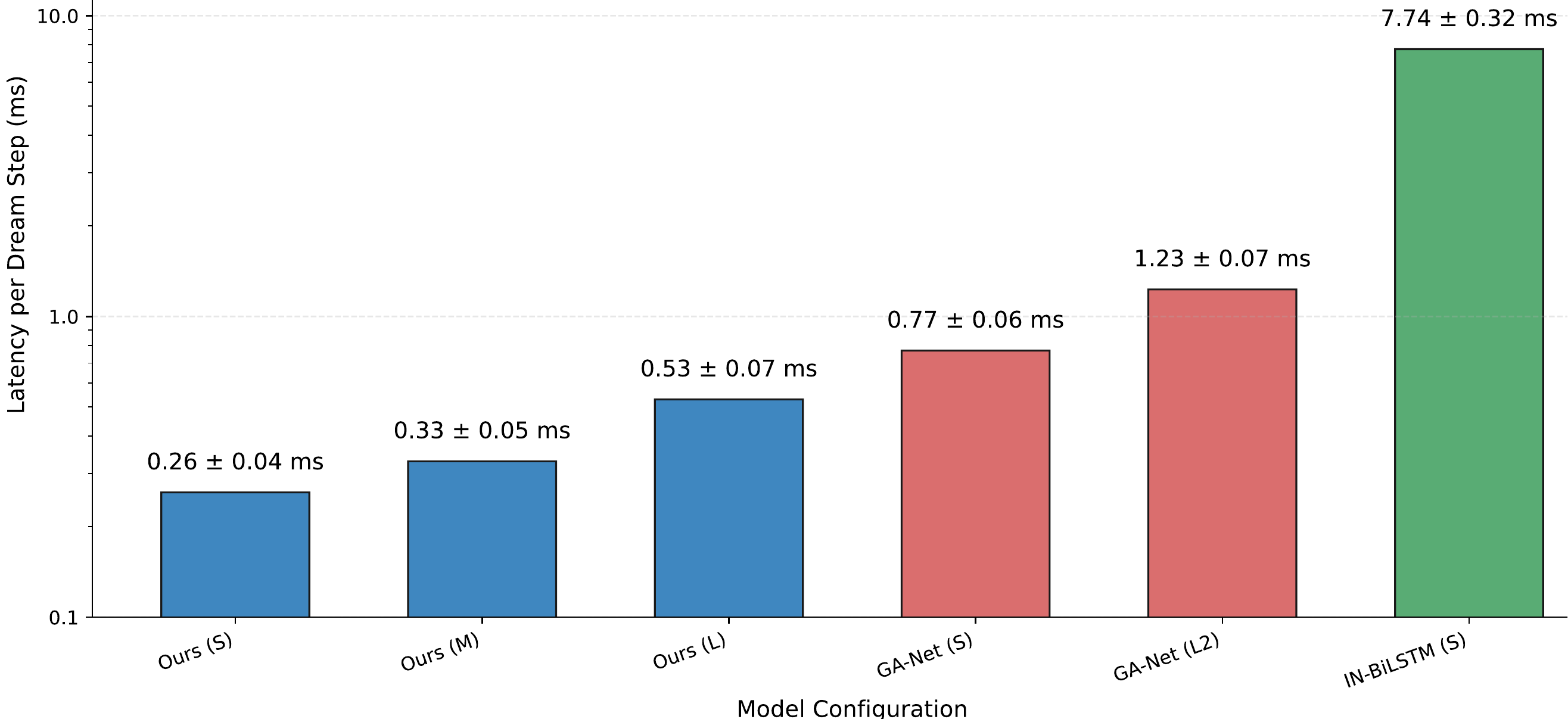}
    \caption{Comparison of mean single-step inference time over different model architectures. S, M and L denote the model sizes as described in Tables \ref{tab:model_sizes} and \ref{tab:model_sizes_ganet}. GA is short for GA-Net. Red lines represent standard deviation. Lower is better. All experiments were conducted on an Nvidia 4060Ti GPU.}
    \label{fig:inference_time}
    \vspace{-0.5cm}
\end{figure}

Finally, we report inference latencies of our models and the best-performing baseline models in Fig. \ref{fig:inference_time} (log-scaled). Despite its higher parameter count, our large model achieves a \textbf{31.17\%} reduction in computation time per prediction step compared to the small GA-Net ($0.53$ ms vs. $0.77$ ms). Our smaller models widen this gap further at the cost of accuracy. IN-BiLSTM (S) shows inference times unsuitable for practical usage, with the larger models taking notably longer to perform a single prediction step. 
The superior inference speed of our architecture stems from projecting the state sequence into a compact latent space. Unlike GNN-based models, which must explicitly compute pairwise edge interactions and decode to physical positions at every time step, our RSSM performs temporal rollouts entirely within the latent space using lightweight recurrent updates. This allows the model to bypass the heavy graph construction overhead during multi-step forecasting. Consequently, our approach drastically accelerates the massively parallel trajectory sampling required for Model Predictive Control.

\subsection{Topology Error}

\begin{figure}[t]
    \centering
    \includegraphics[width=0.47\textwidth]{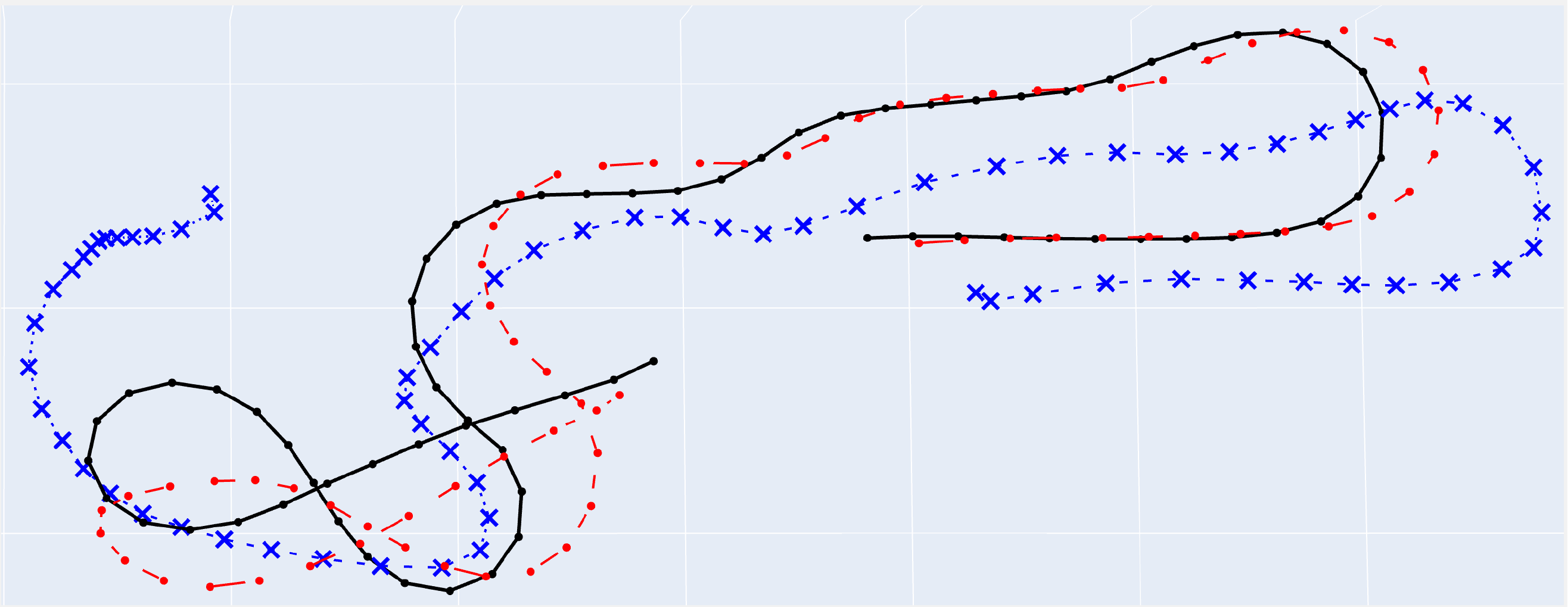}
    \caption{Example of a loop and prediction at $t=20$ of our small model (red) and GA-Net S (blue) compared against the ground truth (black). Our model effectively identifies the loop and accurately reconstructs the rope. GA-Net fails to maintain DLO topology and predicts physically unrealistic link lengths.}
    \label{fig:dlo_loop}
    \vspace{-0.5cm}
\end{figure}

To evaluate the topological fidelity of the learned dynamics, we conduct a quantitative analysis of the models' ability to maintain physical consistency during self-intersections. To automate this, we represent the DLO's topology using Gauss Codes. This numerical form models the DLO as a directed curve and assigns a signed integer to each crossing: positive for an over-crossing ($+i$) and negative for an under-crossing ($-i$) \cite{kauffman_unknotting_2016}. For example, the configuration in Fig. \ref{fig:dlo_loop} yields a Gauss code of $[1, 2, -1, -2]$ when traversed from the right terminus to the left. By comparing predicted and ground-truth codes at each step, we measure the model's preservation of the DLO's structural identity. A match in Gauss codes implies topological equivalence.

\begin{figure}[t]
    \centering
    \includegraphics[width=\columnwidth]{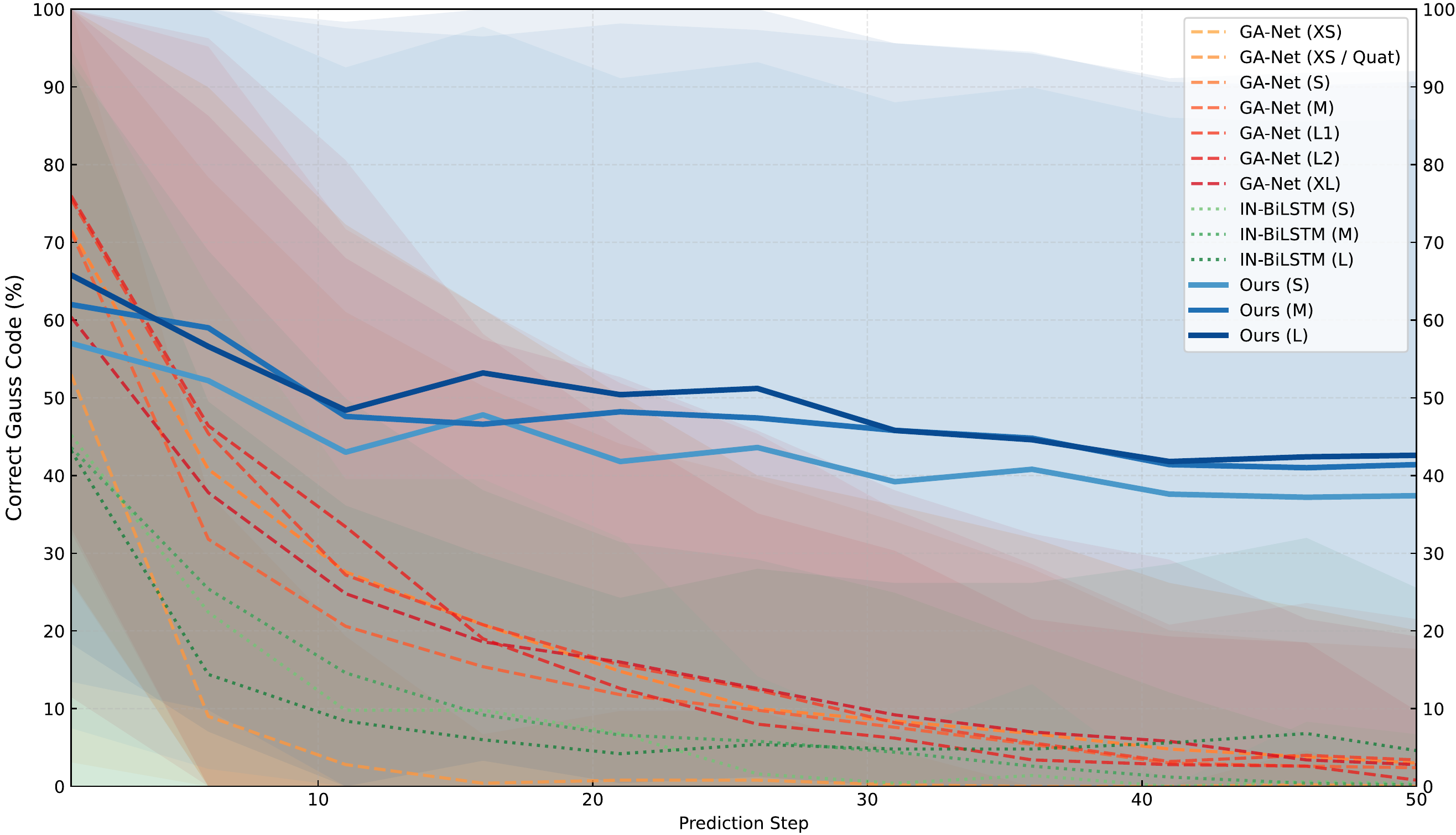}
    \caption{Comparison of Topology Accuracy. Percentage denotes the amount of Gauss Codes exactly matching the ground truth. Higher is better. }
    \label{fig:topology}
\end{figure}

Following a 5-step warmup phase, each model was evaluated using open-loop rollouts across 50-step horizons over 500 unique trajectories. The mean performance and corresponding standard deviation are illustrated in Fig. \ref{fig:topology}. 

Our proposed architecture demonstrates a high degree of topological stability, maintaining mean success rates between 65\% and 38\% throughout the full prediction horizon regardless of model size. In contrast, the baseline models exhibit a precipitous decline in accuracy. The most competitive GA-Net baseline drops to 40\% accuracy by step 10, with all baseline methods falling below the 10\% mark by step 30. Notably, the GA-Net model using our quaternion representation (XS / Quat) drops to 0\% accuracy by step 15. This suggests that while baselines may capture local geometric features in short-term prediction, they fail to preserve the structural identity of the DLO. In contrast, our model maintains topological accuracy in short- and long-horizon prediction.

\section{Conclusion and Future Work}
\label{sec:conclusion}

In this paper, we introduced \textit{RopeDreamer}, a latent dynamics framework that leverages the synergy between Recurrent State Space Models and a Quaternionic Kinematic representation for the manipulation of DLOs. Our experiments demonstrate that by shifting dynamics modeling from local Cartesian graphs to a global latent manifold, we can effectively mitigate the information washout and temporal drift common in existing state-of-the-art baselines. While the latent bottleneck necessitates an initial reconstruction trade-off, our approach achieves a \textbf{40.52\% reduction} in RMSE over 50-step horizons in complex deformation tasks. Crucially, our approach further delivers a substantial increase in topological integrity. Combined with a \textbf{31.17\% improvement} in inference efficiency, this framework provides a robust and scalable backbone for real-time Model Predictive Control in high-dimensional manipulation of DLOs.

Despite these gains, several research avenues remain open. We intend to explore hierarchical latent architectures to further reduce the initial reconstruction error. Another key objective is closing the sim-to-real gap through the integration of online system identification to adapt the latent dynamics to varying material properties, such as cable stiffness or surface friction. Finally, we aim to close the loop by deploying \textit{RopeDreamer} as a predictive primitive within a reinforcement learning framework for various rope manipulation tasks, validating its performance on physical robotic platforms using marker-free vision-based tracking.








\bibliographystyle{ieeetr}
\bibliography{references}

\end{document}